\documentclass[runningheads]{llncs}

\usepackage{accv}

\usepackage{accvabbrv}

\usepackage{graphicx}
\usepackage{booktabs}
\usepackage{amsmath}
\usepackage{amssymb}
\usepackage{multirow}
\usepackage{caption}
\usepackage{algorithm}
\usepackage{algpseudocode}

\usepackage{adjustbox}
\expandafter\def\csname ver@subfig.sty\endcsname{}
\usepackage{subfig}

\usepackage{enumitem}

\usepackage[accsupp]{axessibility}  

\usepackage{hyperref}

\usepackage{orcidlink}

\begin{document}

\title{MeshGS: Adaptive Mesh-Aligned Gaussian Splatting for High-Quality Rendering}

\titlerunning{Adaptive Mesh-Aligned Gaussian Splatting for High-Quality Rendering}



\author{Jaehoon Choi\inst{1} \and
Yonghan Lee\inst{1} \and
Hyungtae Lee\inst{3} \and \\
Heesung Kwon\inst{2} \and  
Dinesh Manocha\inst{1} 
}

\authorrunning{Jaehoon et al.}


\institute{University of Maryland, College Park, USA \and 
DEVCOM Army Research Laboratory, Adelphi, USA\and
BlueHalo, Rockville, USA \\
\email{\{kevchoi,lyhan12\}@umd.edu, hyungtae.lee@bluehalo.com, heesung.kwon.civ@army.mil, dm@cs.umd.edu}}

\maketitle

\begin{abstract}

Recently, 3D Gaussian splatting has gained attention for its capability to generate high-fidelity rendering results. At the same time, most applications such as games, animation, and AR/VR use mesh-based representations to represent and render 3D scenes.  
We propose a novel approach that integrates mesh representation with 3D Gaussian splats to perform high-quality rendering of reconstructed real-world scenes. In particular, we introduce a distance-based Gaussian splatting technique to align the Gaussian splats with the mesh surface and remove redundant Gaussian splats that do not contribute to the rendering. We consider the distance between each Gaussian splat and the mesh surface to distinguish between tightly-bound and loosely-bound Gaussian splats. The tightly-bound splats are flattened and aligned well with the mesh geometry. 
The loosely-bound Gaussian splats are used to account for the artifacts in reconstructed 3D meshes in terms of rendering.
We present a training strategy of binding Gaussian splats to the mesh geometry, and take into account both types of splats. In this context, we introduce several regularization techniques aimed at precisely aligning tightly-bound Gaussian splats with the mesh surface during the training process.
We validate the effectiveness of our method on large and unbounded scene from mip-NeRF 360 and Deep Blending datasets. Our method surpasses recent mesh-based neural rendering techniques by achieving a 2dB higher PSNR, and outperforms mesh-based Gaussian splatting methods by 1.3 dB PSNR, particularly on the outdoor mip-NeRF 360 dataset, demonstrating better rendering quality. We provide analyses for each type of Gaussian splat and achieve a reduction in the number of Gaussian splats by 30\% compared to the original 3D Gaussian splatting.     

\keywords{Neural rendering \and Gaussian splatting \and Mesh}
\end{abstract}

\section{Introduction}
\label{sec:intro}
Rendering realistic depictions of real-world objects and environment has long been a significant challenge with numerous practical applications in fields like computer vision, graphics, and AR/VR. Most graphics and game engines use polygonal mesh-based representations of the objects in the scene \cite{karis2013real}, including real-time rendering and animation. 
Recently, Neural Radiance Fields (NeRF) \cite{mildenhall2021nerf,barron2021mip,barron2022mip,muller2022instant} have demonstrated promising capabilities for 3D reconstruction and novel-view synthesis. However, NeRFs, which rely on volumetric rendering, suffer from slow rendering speeds and are not easily compatible with modern graphics engines \cite{tang2022nerf2mesh,VMesh,Choi2024LTM}. Instead, researchers have been investigating methods based on neural implicit representation to facilitate mesh-based rendering. MobileNeRF \cite{chen2023mobilenerf} utilizes a classical rasterization pipeline, incorporating z-buffers and fragment shaders, to train neural fields based on polygonal mesh representations. Subsequently, various methods \cite{yariv2023bakedsdf,tang2022nerf2mesh,NeRFMeshing,NeuRas,DNMP} first reconstruct a 3D mesh and then utilize this mesh to train appearance models employing high-speed rasterization \cite{opengl,nvdiffrast,PyTorch3D}. These techniques facilitate real-time rendering and demonstrate potential for integration into graphics and game engines. However, all of these methods are constrained by the quality of the mesh surface, particularly exhibiting a loss in rendering quality when dealing with highly detailed structures, a well-known limitation of mesh-based approaches \cite{kazhdan2013screened,wang2021neus}. Given the inherent difficulty in achieving perfect geometric reconstruction for real-world unbounded scenes, rendering algorithms must demonstrate robustness in handling mesh artifacts such as thin-level structures.


More recently, the 3D Gaussian Splatting method \cite{kerbl20233d} has attracted considerable interest due to its high-quality rendering and real-time rendering speed. Compared to mesh-based representation, 3D Gaussian splats excel in capturing intricate details of the scene with high fidelity due to their high degree of freedom in splat positions and shapes. However, 3D Gaussian splats are limited in their ability to represent the geometry of a scene and are not widely applicable to various applications.
SuGaR \cite{guedon2023sugar} extracts mesh solely based on 3D Gaussian splats. Their method shows detailed structure at the object level and tightly binds the 3D Gaussian splats to the surface of the mesh. Other previous works adopt a similar method for tightly binding the 3D Gaussian splat to the mesh triangles \cite{waczynska2024games,qian2023gaussianavatars}.  
This is very useful for rigging, animation, or any deformation tasks. However, this assumes that the mesh can cover nearly all geometrical elements of the target object, such as the head or a human avatar. In unbounded scenes, since meshes frequently exhibit artifacts, we need better strategies to effectively bind 3D Gaussian splats to the mesh surface. 


\textbf{Main Results:} In this paper, we present a new approach to integrate the strengths of mesh and 3D Gaussian splats in rendering and geometry for unbounded large-scale scenes. For geometry reconstruction, we first employ the BakedSDF \cite{yariv2023bakedsdf} technique and extract a 3D mesh using Marching cube algorithm \cite{marchingcube}. In unbounded large-scale scenes, meshes typically consist of a large number of triangles to represent complex geometric objects in the scene. Properly initializing Gaussian splats with millions of vertices and triangles becomes exceptionally challenging. Inspired by LTM \cite{Choi2024LTM}, we employ mesh decimation and eliminate redundant triangles to extract a lightweight mesh with fewer triangles. Based on this lightweight mesh, we initialize the Gaussian splats for each mesh triangle. Subsequently, we introduce a mesh-based Gaussian splatting method that consists of two components. First, we conduct forward splatting considering mesh geometry and eliminate any Gaussian splats that are occluded by the mesh surface. This impacts our training process to align the Gaussian splats with the explicit mesh geometry and to remove redundant Gaussian splats positioned behind the mesh surface considering all training viewpoints. 

Secondly, we segregate tightly-bound Gaussian splats from loosely-bound Gaussian splats using the distance between each Guassian splat's center and their corresponding mesh triangles. As current mesh reconstruction methods cannot fully capture the 3D geometry of real-world scenes, we introduce loosely-bound Gaussian splats to depict regions where explicit mesh-based representation is not available, thus representing their appearance. We employ different training and densification strategies for each type of splat. The tightly-bound Gaussian splats are aligned with the mesh surface through various geometric regularization techniques, whereas the optimization of loosely-bound Gaussian splats relies solely on image supervision. 

Our main contributions include:

\begin{itemize}
    \item We present a novel approach for mesh-based Gaussian splatting. We integrate Gaussian splats with the triangle mesh and introduce a new method for initializing and training Gaussian splats based on the triangle mesh. 
    \item We differentiate between two types of Gaussian splats based on their distance from the mesh surface. Subsequently, we introduce different training strategies for each type of Gaussian splat to accommodate different rendering requirements.
    \item Our method surpasses the rendering quality of the state-of-the-art mesh-based neural rendering algorithms \cite{yariv2023bakedsdf,Choi2024LTM} by 2dB PSNR on mip-NeRF360 dataset. Compared to the original 3D Gaussian splatting \cite{kerbl20233d}, we achieve comparable rendering quality while utilizing 30\% fewer Gaussian splats for rendering. 
\end{itemize}

\section{Related Work}
\label{sec:related}
\noindent\textbf{Neural Rendering.}
Neural Rendering has recently received considerable attention. Neural Radiance Fields (NeRF) \cite{mildenhall2021nerf} utilize a multi-layer perceptron (MLP) to encode a scene by applying differentiable volumetric rendering to represent RGB radiance and density. Various methods have been proposed to accelerate the training time and rendering speed by leveraging data structures \cite{fridovich2022plenoxels,chen2022tensorf,sun2022direct}, encoding techniques \cite{muller2022instant,barron2021mip,barron2022mip}, and baking processes \cite{SNeRG,MeRF}. Other researchers utilize neural implicit surface representation \cite{eikonal}, employing MLP to map coordinates to a signed distance function (SDF) or occupancy grids. These approaches \cite{yariv2021volsdf,wang2021neus,oechsle2021unisurf,yu2022monosdf,choi2023tmo,yariv2023bakedsdf,li2023neuralangelo} involve converting volume density to SDF or occupancy and perform training with differentiable volumetric rendering. 


\noindent\textbf{Mesh-Based Rendering.}
Polygonal mesh-based rendering \cite{opengl} is a classic problem in computer graphics to visualize 3D models on a 2D screen using rasterization. To address rendering speed issue, recent neural rendering methods also leverage explicit mesh and bake the neural appearance, enabling mesh-based neural rendering. SNeRG \cite{SNeRG} builds on deferred shading \cite{deering1988triangle} and bakes neural appearance into sparse voxel grid. MobileNeRF \cite{chen2023mobilenerf} directly optimizes the triangle faces and embeds opacities and features into texture maps through a baking process.   
Subsequently, other approaches involve baking the neural fields into textures and representing view-dependent effects using either a small neural network \cite{NeRFMeshing,tang2022nerf2mesh,NeuRas,DNMP} or spherical Gaussians \cite{yariv2023bakedsdf}. VMesh \cite{VMesh} proposes hybrid representation combining mesh and volume rendering to achive high efficiency and expressiveness. Shells \cite{AdaptiveShell} extract the mesh and conduct volumetric rendering within a narrow band surrounding the mesh. LTM \cite{Choi2024LTM} employs significant mesh decimation to achieve efficient mesh representation for large-scale scenes while simultaneously training neural appearance and geometry through differentiable rendering \cite{nvdiffrast}. 
Many existing methods encounter limitations due to mesh artifacts stemming from 3D reconstruction techniques in real-world scenarios, posing challenges in achieving photorealistic rendering quality with neural fields. Our approach aims to address this limitation by leveraging Gaussian splatting for mesh-based rendering.


\noindent\textbf{Point-Based Rendering and Gaussian Splatting}
Point-based rendering is a technique for rendering 3D scenes using point-based geometry representation instead of traditional polygon meshes. In this context, a variety of recent approaches have been proposed for the point-based rendering \cite{kopanas2021point,ruckert2022adop}, sphere rendering \cite{lassner2021pulsar}, and differentiable splatting \cite{yifan2019diffsplat}. The 3D Gaussian Splatting \cite{kerbl20233d} employs 3D Gaussian splats \cite{zwicker2002ewa}, which are rendered through sorting and rasterization processes. SuGaR \cite{guedon2023sugar} utilizes Poisson reconstruction \cite{kazhdan2006poisson} to extract a mesh from 3D Gaussian splats and bind them to the triangle for rendering. GaussianAvatars \cite{qian2023gaussianavatars} and GaMeS \cite{waczynska2024games} focus on tightly binding 3D Gaussian splats to meshes and rendering dynamic scenes, including human avatars or single objects.  However, these methods are mainly tailored for a single object or human avatars. Our goal is to develop a new mesh-based Gaussian splatting technique suitable for large-scale scenes, even in cases where the mesh only captures moderate levels of geometry. More recently, 2D Gaussian splatting \cite{huang20242d} utilize the 2D Gaussian primitives instead of 3D Gaussian splats. In our paper, we focus solely on 3D Gaussian splat representation and the integration of 3D Gaussian splats with the mesh. We provide a more detailed analysis comparing 2DGS \cite{huang20242d} with our method in the supplementary material.


\section{Our Method}
\label{sec:method}

\begin{figure}[t]
    \centering
    \includegraphics[width=0.92\linewidth]{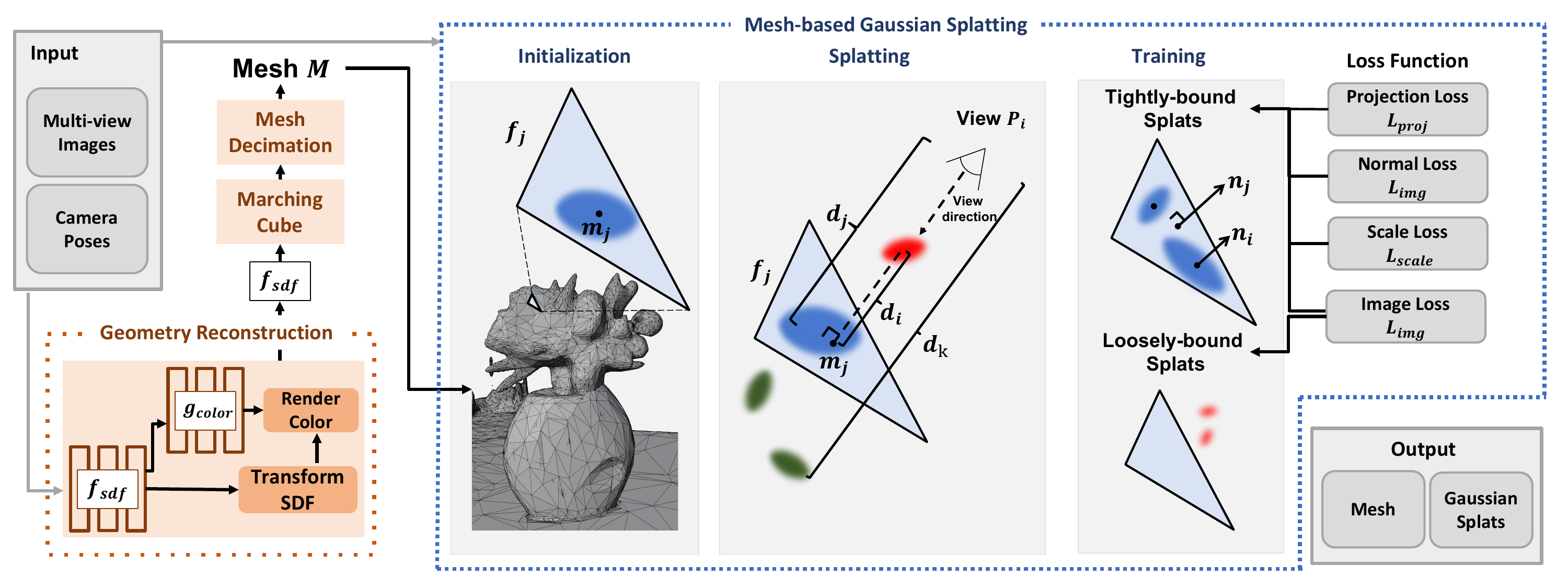}
    \caption{
    \textbf{Our Approach:} In Section \ref{sec:GeometryReconstruction}, we jointly train the geometry $f_{sdf}$  and appearance $g_{color}$ using differentiable volumetric rendering (shown in orange). Then, we extract the lightweight mesh $M$ which is important for the initialization of Gaussian splats. Next, in Section \ref{sec:Mesh-based Gaussian Splatting}., we present the mesh-based Gaussian splatting for training, which involves the removal of Gaussian splats (green Gaussian splats) occluded by mesh surface. Given the distance $d_{i}$ between each Gaussian splat and its corresponding triangle, we distinguish between tightly-bound Gaussian splats (blue splat) and loosely Gaussian splats (red splat). In Section \ref{sec:Training}, we introduce the training and densification strategy for both tightly-bound Gaussian splats and loosely Gaussian splats.}
    \label{fig:framework}
    \vspace{-3mm}
\end{figure}

Given posed images, our goal is to reconstruct both the precise geometry and rendering high-fidelity images of unbounded large-scale scenes. An overview of our pipeline is illustrated in Fig. \ref{fig:framework}. First, we employ the BakedSDF \cite{yariv2023bakedsdf} technique and extract a 3D mesh $M$ via Marching Cubes \cite{marchingcube}. Following the method in \cite{Choi2024LTM}, we significantly reduce the number of triangles in the mesh to facilitate the integration with Gaussian splats to generate high-quality rendering. Next, we bind the Gaussian splats $G$ to the explicit mesh to represent their appearance.  

\subsection{Geometry Reconstruction}
\label{sec:GeometryReconstruction}
During this stage, we leverage the neural surface representation to extract a mesh and decimate it for integrating the Gaussian splats with the mesh. The original 3D Gaussian Splatting method (3DGS) \cite{kerbl20233d} is specifically tailored to sparse point clouds generated by Structure-from-Motion (SfM) techniques \cite{schonberger2016structure}.
Dense point clouds are not conducive to rendering and can even be detrimental to the rendering quality. Hence, it is imperative to meticulously design mesh reconstructions for the initialization of Gaussian splats. Following the approach of BakeSDF \cite{yariv2023bakedsdf}, we jointly train two multilayer perceptrons (MLPs) to encode both the signed distance field $f_{sdf}$ and the appearance field $g_{color}$, as shown in the \textit{Geometry Reconstruction} block in Fig. \ref{fig:framework}. Both MLPs are optimized using differentiable volumetric rendering and Eikonal regularization techniques\cite{eikonal}. Details are described in \cite{yariv2023bakedsdf}. Then, we employ Marching Cube algorithm \cite{marchingcube} to obtain a mesh $M = \{V, F\}$ with vertices $V$ and faces $F$ extracted from $f_{sdf}$. 

\begin{figure}[t]
    \centering
    \includegraphics[width=0.99\linewidth]{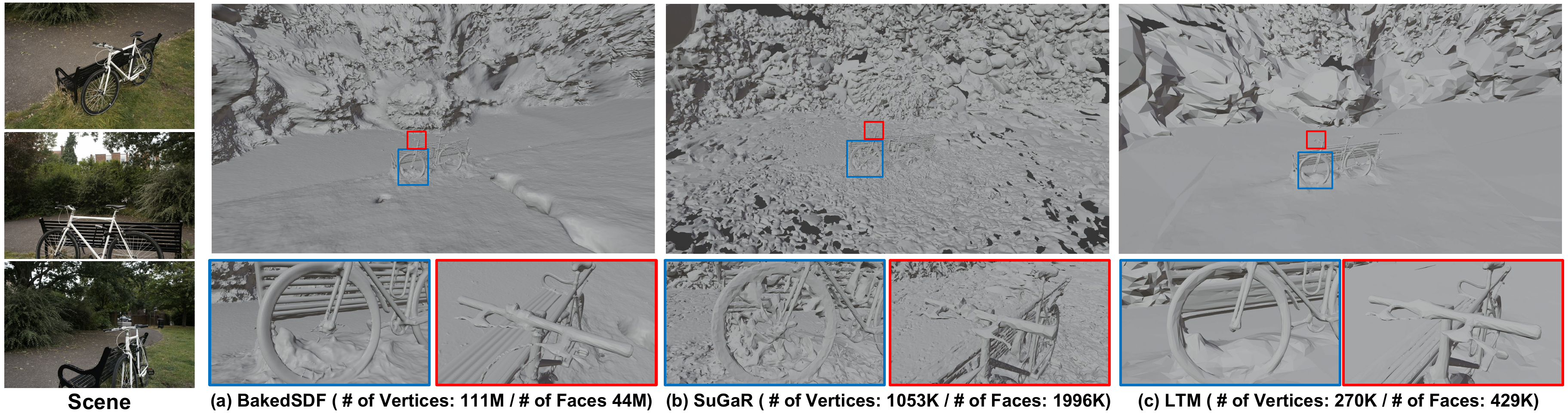}
    \caption{The examples of extracted mesh from BakedSDF \cite{yariv2023bakedsdf}, SuGaR \cite{guedon2023sugar}, and LTM \cite{Choi2024LTM}. We employ LTM \cite{Choi2024LTM} to extract a lightweight mesh with moderate triangle count, using it for the initialization of Gaussian splats. The primary reason is that LTM utilizes a minimal number of triangles while maintaining geometric quality. Zoomed-in regions visualize that the mesh frequently lacks highly detailed geometry or complex structures. ``K" and ``M" denotes the $10^3$ and $10^6$ units, respectively.
    }
    \label{fig:mesh_comparison}
\end{figure}

However, in large-scale scenes, the output mesh occupies a significant number of triangles (average 27 millions) \cite{yariv2023bakedsdf}. Given occupying a large amount of triangles, they still cannot represent detailed geometry or reflective surfaces due to the strong smoothness regularization. Furthermore,  a large number of triangles are allocated to distorted regions, especially in background regions. Several prior mesh-based Gaussian splatting methods \cite{qian2023gaussianavatars,guedon2023sugar,waczynska2024games} assume that at least one Gaussian splat is attached to the mesh surface to represent the appearance of its triangles. 
If we strictly follow their strategy, we allocate redundant Gaussian splats to distorted regions, resulting in a rapid increase in GPU memory usage after just a few training and densification steps. In Fig.\ref{fig:mesh_comparison} (a), the mesh reconstructed by BakedSDF occupies 44 million triangles to represent geometry. The original 3DGS \cite{kerbl20233d} necessitates 4 million Gaussian splats to achieve high-fidelity rendering quality, yet the very large number of Gaussian splats undermines rendering quality. SuGaR also imposes constraints on the vertex count (10K$\sim$15K); however, a significant number of vertices are needed to faithfully represent the geometry, particularly in expansive, unbounded scenes. We observe that SuGaR sacrifices geometric quality, particularly in background regions (see Fig.\ref{fig:mesh_comparison} (b)), and then trains Gaussian splats attached to mesh faces. However, since we aim to achieve moderate geometry quality, adopt the methodology of LTM \cite{Choi2024LTM}. They utilize aggressive mesh decimation based on the Quadric Error Metric \cite{garland1997surface,garland1999quadric} to efficiently represent moderate geometry with a fewer number of triangles  ( see Fig. \ref{fig:mesh_comparison} (c)). Using this mesh as a foundation, we can easily integrate Gaussian splatting with the explicit mesh.

\subsection{Mesh-based Gaussian Splatting}
\label{sec:Mesh-based Gaussian Splatting}
\noindent\textbf{3D Gaussian Splatting: }
3D Gaussian Splatting (3D-GS) \cite{kerbl20233d} employs millions of 3D Gaussians $G$ to represent a 3D scene. These 3D Gaussian splats are constructed based on a sparse point cloud obtained from Structure from Motion \cite{schonberger2016structure}, incorporating various attributes to model both the geometry and appearance of the scene. The geometry attributes of each 3D Gaussian (ellipsoids) $G_{i}$ is defined by its center (position) $p_{i} \in \mathbb{R}^3$ and 3D covariance matrix $\Sigma_{i} \in \mathbb{R}^{3\text{x}3}$ in world space:    
\begin{equation} \label{eq:3DGaussians}
G_{i}(x) = e^{-\frac{1}{2}(x-p_{i})^T \Sigma_{i}^{-1}(x-p_{i})}.
\end{equation}
 To constrain valid covariance matrices, 3D-GS utilizes a semi-definite parameterization $\Sigma_{i} = R_{i}S_{i}{S_{i}}^{T}{R_{i}}^{T}$ with two learnable components: a scaling matrix $S_{i} \in \mathbb{R}^{3}$ and a rotation matrix $R_{i} \in \mathbb{R}^{3\text{x}3}$, encoded by a quaternion $q \in \mathbb{R}^4$. For rendering, given a viewpoint defined by rotation $R_{cw} \in \mathbb{R}^{3\text{x}3}$ and translation $t_{cw} \in \mathbb{R}^{3}$, the 3D Gaussians are projected into the camera space:
$p^{c}_{i} = R_{cw}p_{i} + t_{cw}$  and $\Sigma^{c}_{i} = R_{cw}\Sigma_{i}{R_{cw}}^{T}$. Next, to transform camera space into ray space, 3D-GS utilize the affine transformation $J_{i} \in \mathbb{R}^{2\text{x}3}$ defined by $p^{c}_{i}$ following \cite{zwicker2002ewa}, and the 2D covariance matrix in ray space can then be defined as $\Sigma'_{i} = J_{i}\Sigma^{c}_{i}{J_{i}}^{T}$. We can obtain the 2D Gaussian $g_{i}$ defined by corresponding 2D covariance $\Sigma'_{i}$ and 2D center $p'_{i}$ as follows:
\begin{equation} \label{eq:32dGaussians}
g_{i}(x) = e^{-\frac{1}{2}(x-p'_{i})^T {\Sigma'_{i}}^{-1}(x-p'_{i})}.
\end{equation}

Additionally, each Gaussian includes an per-point opacity $o_{i} \in [0, 1]$ and the view-dependent color $c_{i}$ represented by spherical harmonic (SH) \cite{kerbl20233d}. 3D-GS orders all the Gaussians that contributes to a pixel and renders 2D image via alpha blending with the sorted Gaussians using the following equation:
\begin{equation} \label{eq:3DGSrendering}
c = \sum_{i=1}^{n} c_{i}\alpha_{i}\prod_{j=1}^{i-1}(1-\alpha_{j}), \:\:\:\: \alpha_{i} = o_{i}g_{i}(x).
\end{equation}
The blending weight $\alpha_{i}$ is obtained by computing the 2D projection of the 3D Gaussian $g_{i}(x)$ multiplied by a per-point opacity $o_{i}$.

\noindent\textbf{Distance-based Gaussian Splatting:}
In this section, we present our approach that builds the connection between large-scale mesh and Gaussian splats. Before training, we initialize a 3D Gaussian splats at the center $m_{j}$ of each triangles $f_{j}$. Given mesh $M$, we can render a depth $D_{i}$ from a given viewpoint $P_{i}$. When rendering for each viewpoint, we utilize the corresponding viewpoint depth map to mask out the Gaussian splats occluded by the surface. In light of our assumption that the mesh surface is entirely opaque, Gaussian splats occluded by it should be excluded for rendering purposes. In the \textit{Splatting} block shown in Fig. \ref{fig:framework}, given that the depth $d_{k}$ of the green Gaussian splats is greater than that of $d_{j}$, those Gaussian splats located behind the mesh surface do not contribute to rendering pixels. Furthermore, during the training process, we assess the visibility of Gaussian splats across all training views and identify those located behind the mesh surface hat are never observed by any of the training views. Then, we remove all redundant Gaussian splats situated behind the mesh surface at specific training iterations.

Given that reconstructed meshes in real-world environments often exhibit geometric artifacts, our goal is to render not only well-represented regions but also those inaccuracies resulting from severe mesh artifacts. When we tightly bind all Gaussian splats exclusively to the mesh surface, rendering performance suffers, particularly when addressing severe mesh artifacts. Therefore, during the training step, We calculate the distance $d_{i}$ between each Gaussian splat's center $p_{i}$ and corresponding mesh faces $f_{j}$. We treat tightly-bound Gaussian splats $G^{t}_{i} \:\: (d_{i} < d_{th})$ and loosely-bound Gaussian splats $G^{l}_{i} \:\: (d_{i} > d_{th})$, distinctly.
If the Gaussian splats are located at a distance greater than a certain threshold $d_{th}$ from the mesh surface, we classify them as loosely-bound Gaussian splats. We aim to flatten $G^{t}_{i}$ and align these splats with the mesh surface. 


\subsection{Training for Tightly-bound and Loosely-bound Gaussian Splats}
\label{sec:Training}
\noindent\textbf{Objective Function:} Following 3D-GS \cite{kerbl20233d}, we employ the combination of $L_{1}$ loss and D-SSIM as the image loss $L_{img}$ between the rendered image $\hat{I}_{i}$ and the ground truth image $I_{i}$ expressed as follows:
\begin{equation} \label{eq:imageloss}
    L_{img} = (1-\lambda_{img})L_{1}(\hat{I}_{i},I_{i}) + \lambda_{img}SSIM(\hat{I}_{i}, I_{i})).
\end{equation}

We utilize the normal consistency loss $L_{nc}$ to refine tightly-bound Gaussian splats $G^{t}$ to align with the corresponding mesh surface. To estimate normals $n_{i}$ for Gaussian splats, we first identify the minimum scale axis among the scale parameters $S_{i}$, and then apply the rotation matrix $R_{i}$ to transform them into world coordinates. Then, we exploit the cosine similarity between the splat's normals $n_{i}$ and the corresponding face normals $n_{f}$ as follows:
\begin{equation} \label{eq:ncloss}
    L_{nc} = \sum_{i \in G^{t},f \in F} (1 - n_{i} \cdot n_{f}).
\end{equation}

Furthermore, we apply the regularization to suppress the scale parameters $S_{i} = (s_{1}, s_{2}, s_{3}) \in \mathbb{R}^{3}$ of tightly-bound Gaussian splats $G^{t}$ and align them with the mesh surface. We regularize the minimum scale and then the maximum scale with threshold value $\rho$.  
\begin{equation} \label{eq:scaleloss}
L_{scale} = \sum_{i \in G^{t}}(\lambda_{min}|\text{min}(s_{1}, s_{2}, s_{3})| + \lambda_{max}|\text{max}(s_{1}, s_{2}, s_{3}) - \rho|).
\end{equation}
This ensures that the tightly-bound Gaussian splats do not expand excessively in comparison to the corresponding mesh face $f_{j}$, thus hindering the rendering process. Finally, given the mesh $M$ consisting of face $f_{j}$ and its face normal $n_{j}$, we project the Gaussian Splat's center $P_{i}$ to the closest point $s_{i}$ on the mesh surface. 
\begin{equation} \label{eq:projectionloss}
\begin{split}
     &s_{i} = P_{i} + ((P_{i} - v_{i}) \cdot f_{j}) \cdot f_{j}, \\ 
     &L_{proj} =  \sum_{i \in G^{t}}\parallel s_{i} - P_{i} \parallel_{2}. 
\end{split}
\end{equation}
This loss function ensures that the tightly-bound Gaussian splats remain close to the mesh surface, preventing them from moving too far from the surface. The total loss function $L$ is defined as: 

\begin{equation} \label{eq:totalloss}
    L = L_{img} + \lambda_{nc} L_{nc} + L_{scale} +\lambda_{proj} L_{proj}.
\end{equation}

where $\lambda$ denotes the weight used to balance between the different loss terms.
\noindent\textbf{Densification Strategy}
Here, we adopt two different densification strategies for loosely-bound Gaussian splats $G^{l}$ and tightly-bound Gaussian splats $G^{t}$. Regarding $G^{l}$, we follow the method outlined in the original 3DGS \cite{kerbl20233d}. The original Gaussian densification strategy inherently overlooks the existing 3D geometry. In terms of $G^{t}$, we find the mesh triangle close to $G^{t}$ and generate new Gaussian splats by employing random sampling from the corresponding mesh face. For both cloning and splitting, we conduct point sampling from the triangle and utilize this position to initialize new Gaussian splats. During cloning, additional attributes are copied to the new Gaussians, while during splitting, their scale is divided by a factor of 1.6. 

\section{Experiments}
\label{sec:expreiments}

\subsection{Experiment Settings}
\label{sec:expreiment_settings}
\noindent\textbf{Datasets:} For comparison, we assess the rendering quality of our method using the extensively utilized real-world large-scale scenes published by Mip-NeRF360 \cite{barron2022mip}. Furthermore, to assess the generalization capability, we employ a different large-scale dataset which are two scenes from Deep Blending \cite{hedman2018deep}.


\noindent\textbf{Implementation Details:}
For geometry reconstruction section in Sec \ref{sec:GeometryReconstruction}, we utilize the LTM \cite{Choi2024LTM} code for extracting mesh from neural field and decimating those mesh. In Sec \ref{sec:Mesh-based Gaussian Splatting}, when applying distance-based Guassian splatting, we add the 0.01 to the depth $D_{i}$. We prune the Gaussian splats located behind the mesh at every 500 training iterations. To distinguish the tightly-bound and loosely-bound Gaussian splats, we set the distance threshold $d_{th}$ to 0.01. For training Gaussian splatting, we generally follow the original 3DGS paper \cite{kerbl20233d}.  we run all experiments for 30k iterations.  We set the loss weights to $\lambda_{img}$=0.001, $\lambda_{nc}$=0.1, $\lambda_{min}$=0.1, $\lambda_{max}$=10, and $\lambda_{proj}$=50. We set the threshold $\rho$ to 0.1. Following the train/test split in Mip-NeRF360, we subsampled every eight frames for test sets. We evaluate all our method trained with 30K iterations.  

\subsection{Experimental Results}
We compare our proposed method with three different approaches: LTM \cite{Choi2024LTM}, which represents the state-of-the-art method for reconstructing lightweight meshes and training a neural field for the corresponding mesh. SuGaR \cite{guedon2023sugar}, which builds upon the original 3DGS \cite{kerbl20233d} by reconstructing the mesh and binding the Gaussian splats to the corresponding mesh. 3DGS \cite{kerbl20233d}, which relies on sparse point clouds instead of an explicit mesh representation. As BakedSDF \cite{yariv2023bakedsdf} has not released public code, we include only metrics reported in their original paper.

In Table \ref{table:rendering-psnr-comparion-outdoor}, we report the rendering quality using the standard PSNR, SSIM \cite{wang2004image}, and LPIPS \cite{zhang2018unreasonable} on outdoor scenes in mip-NeRF 360 dataset \cite{barron2022mip}. Generally, mesh-based Gaussian splatting methods, including our method, SuGaR \cite{guedon2023sugar}, and GaMeS \cite{waczynska2024games}, exhibit superior performance (with around 2 PSNR increase) compared to state-of-the-art mesh-based neural rendering techniques such as LTM \cite{Choi2024LTM}, BakedSDF \cite{yariv2023bakedsdf}, and others. Our method also demonstrates better quality compared to other recent mesh-based Gaussian splatting methods such as SuGaR \cite{guedon2023sugar} (with a 1.3 PSNR increase) and GaMeS \cite{waczynska2024games} (with a 0.9 PSNR increase). Table \ref{table:rendering-psnr-comparion-indoor} reports the rendering quality on indoor scenes in mip-NeRF 360 dataset \cite{barron2022mip}. Our method also exhibits higher rendering quality compared to GaMeS \cite{waczynska2024games} (with a 2.0 PSNR increase) and SuGaR \cite{guedon2023sugar} (with a 0.2 PSNR increase). Overall, our method shows slightly lower rendering quality compared to the original 3DGS. Without considering geometry, 3D Gaussian splats are more flexible in representing the scene based solely on image loss. This is why most mesh-based splatting methods are not superior to the original 3DGS. Our goal is to find a balance between rendering quality and geometric alignment. Thus, we emphasize the performance by solely comparing mesh-based neural rendering and mesh-based Gaussian splatting.

\begin{table}[t]
    \centering
    \begin{adjustbox}{width=\linewidth,center}
    \begin{tabular}{l|c|c|cccc}
    \toprule
    \multirow{2}{*}{Method}&\multirow{2}{*}{Mesh}&\multirow{2}{*}{Rendering}&\multicolumn{4}{c}{Outdoor (PSNR \(\uparrow\)  / SSIM \(\uparrow\)  / LPIPS \(\downarrow\)) } \\
    & & & Bicycle & Garden & Stump & Mean \\
    \midrule
    Instant-NGP \cite{muller2022instant} & X & Volume & \:\:22.1 / 0.49 / 0.49\:\:& \:\:24.5 / 0.65 / 0.31\:\: & \:\:23.6 / 0.57 / 0.45\:\: & \:\:23.4 / 0.57 / 0.42\:\: \\
    mip-NeRF 360 \cite{barron2022mip} & X & Volume &  24.4 / 0.68 / 0.30 & 26.9 / 0.81 / 0.17 & 26.4 / 0.74 / 0.26 & 25.9 / 0.75 / 0.24 \\
    3DGS \cite{kerbl20233d} & X & Splat & 25.2 / 0.77 / 0.20 & 27.4 / 0.86 / 0.10 & 26.5 / 0.77 / 0.21 & 26.4 / 0.70 / 0.17 \\
    \midrule
    MobileNeRF \cite{chen2023mobilenerf} & O & Mesh & 21.7 / 0.43 / 0.51  & 18.8 / 0.59 / 0.36 & 23.9 / 0.56 / 0.43 & 21.5 / 0.53 / 0.43 \\
    NeRF2Mesh \cite{tang2022nerf2mesh} & O & Mesh & 22.1 / 0.48 / 0.51 & 23.4 / 0.55 / 0.40 & 22.5 / 0.54 / 0.46 & 22.7 / 0.52 / 0.46 \\
    BakedSDF \cite{yariv2023bakedsdf} & O & Mesh & 22.0 / 0.57 / 0.37 & 24.9 / 0.75 / 0.21 & 23.6 / 0.59 / 0.37 & 23.5 / 0.64 / 0.32 \\
    LTM \cite{Choi2024LTM} & O & Mesh & 22.4 / 0.52 / 0.44 & 23.5 / 0.64 / 0.30 & 23.7 / 0.57 /0.42 & 23.2 / 0.58 / 0.39  \\ 
    SuGaR \cite{guedon2023sugar} & O & Splat & 23.1 / 0.64 / 0.34 & 25.3 / 0.77 / 0.22 & 24.7 / 0.68 / 0.34 & 24.4 / 0.70 / 0.30  \\ 
    GaMeS \cite{waczynska2024games} & O & Splat & 23.4 / 0.67 / 0.33 & 26.3 / 0.83 / 0.14 & 24.6 / 0.66 / 0.35  & 24.8 / 0.72 / 0.39 \\  
    \midrule
    MeshGS (Ours) & O & Splat & \underline{24.4} / \underline{0.71} / \underline{0.26} & \underline{26.8} / \underline{0.85} / \underline{0.12} & \underline{25.8} / \underline{0.74} / \underline{0.25} & \underline{25.7} / \underline{0.77} / \underline{0.21}  \\
    MeshGS* (Ours) & O & Splat & \textbf{24.9}\:/\:\textbf{0.75}\:/\:\textbf{0.25} & \textbf{26.8}\:/\:\textbf{0.86}\:/\:\textbf{0.12} & \textbf{26.3}\:/\:\textbf{0.76}\:/\:\textbf{0.22} & \textbf{26.0}\:/\:\textbf{0.79}\:/\:\textbf{0.20} \\
    \bottomrule
    \end{tabular}
    \end{adjustbox}    
    \vspace{1mm}
    \caption{\textbf{Quantitative Comparison on outdoor mip-NeRF 360 Dataset.}  
    Splat-based rendering demonstrates superior performance compared to mesh-based neural rendering techniques such as LTM \cite{Choi2024LTM} and BakedSDF \cite{yariv2023bakedsdf}. MeshGS* did not utilize any regularization techniques to tightly align Gaussian splats; instead, it solely applied image loss $L_{img}$ for training. Compared to GaMeS \cite{waczynska2024games} and SuGaR \cite{guedon2023sugar}, our method enhances the PSNR by 0.9 dB and 1.3 dB, respectively.
    }
    \vspace{-10mm}
    \label{table:rendering-psnr-comparion-outdoor}
\end{table}

\begin{table}[t]
    \centering
    \begin{adjustbox}{width=\linewidth,center}
    \begin{tabular}{l|c|c|cccccc}
    \toprule
    \multirow{2}{*}{Method}&\multirow{2}{*}{Mesh}&\multirow{2}{*}{Rendering}&\multicolumn{5}{c}{Indoor (PSNR \(\uparrow\)  / SSIM \(\uparrow\)  / LPIPS \(\downarrow\))}\\ 
    & & & Room & Counter & Kitchen & Bonsai & Mean \\
    \midrule
    Instant-NGP \cite{muller2022instant} & X & Volume & \:\:29.2 / 0.85 / 0.30\:\: & \:\:26.43 / 0.80 / 0.34\:\: & \:\:28.5 / 0.82 / 0.25\:\: & \:\:30.3 / 0.89 / 0.23\:\:  & \:\:28.6 / 0.84 / 0.28\:\: \\
    mip-NeRF 360 \cite{barron2022mip} & X & Volume &  31.6 / 0.91 / 0.21 & 29.5 / 0.89 / 0.20 & 32.3 / 0.92 / 0.13 & 33.5 / 0.94 / 0.18 & 31.7 / 0.92 / 0.18 \\
    3DGS \cite{kerbl20233d} & X & Splat & 30.6 / 0.91 / 0.26 & 28.7 / 0.90 / 0.29 & 30.3 / 0.92 / 0.22 & 31.9 / 0.94 / 0.29 & 30.4 / 0.92 / 0.19  \\
    \midrule
    MobileNeRF \cite{chen2023mobilenerf} & O & Mesh & 28.9 / 0.85 / 0.28 & 25.1 / 0.72 / 0.29 & 26.8 / 0.79 / 0.79 & 23.8 / 0.71 / 0.72 & 26.2 / 0.77 / 0.52 \\
    NeRF2Mesh \cite{tang2022nerf2mesh} & O & Mesh & 25.7 / 0.79 / 0.35 & 23.9 / 0.71 / 0.35 & 24.0 / 0.61 / 0.36 & 25.0 / 0.77 / 0.29 & 24.7 / 0.72 / 0.34 \\
    BakedSDF \cite{yariv2023bakedsdf} & O & Mesh & 28.7 / 0.87 / 0.25 & 25.7 / 0.81 / 0.28 & 26.7 / 0.82 / 0.24  & 27.2 / 0.85 / 0.26 & 27.0 / 0.84 / 0.26 \\
    LTM \cite{Choi2024LTM} & O & Mesh & 29.3 / 0.88 / 0.23 & 25.1 / 0.77 / 0.27 & 26.5 / 0.80 / 0.21 & 27.3 / 0.84 / 0.22 & 27.1 / 0.82 / 0.23 \\ 
    SuGaR \cite{guedon2023sugar} & O & Splat  & 30.0 / 0.91 / 0.25  & 27.6 / 0.89 / 0.23 & 29.6 / 0.91 / 0.16  & 30.5 / 0.93 / 0.22  & 29.4 / 0.91 / 0.22 \\ 
    GaMeS \cite{waczynska2024games} & O & Splat & 28.8 / 0.89 / 0.26 & 26.4 / 0.84 / 0.29  & 27.2 / 0.86 / 0.22  & 27.8 / 0.89 / 0.29 & 27.6 / 0.87 / 0.26 \\   
    \midrule
    MeshGS (Ours) & O & Splat & \underline{30.3} / \underline{0.90} / \underline{0.24} & \textbf{28.0} / \underline{0.89} / \textbf{0.21}  & \textbf{29.7}\:/\:\textbf{0.90}\:/\:\textbf{0.17}  & \textbf{30.7}\:/\:\textbf{0.93} / \underline{0.20} & \textbf{29.6}\:/\:\textbf{0.91}\:/\:\textbf{0.20} \\
    MeshGS* (Ours) & O & Splat & \textbf{30.9}\:/\:\textbf{0.93}\:/\:\textbf{0.23} & \underline{27.8} / \textbf{0.90} / \underline{0.21} & \underline{29.3} / \underline{0.90} / \underline{0.17} & \underline{30.5} / \underline{0.93} / \textbf{0.19} & 29.6 / 0.91 / 0.20  \\
    \bottomrule
    \end{tabular}
    \end{adjustbox}
    \vspace{1mm}
    \caption{\textbf{Quantitative Comparison on indoor mip-NeRF 360 Dataset.} Splat-based rendering exhibits superior performance when compared to mesh-based neural rendering techniques.  MeshGS* did not utilize any regularization techniques to tightly align Gaussian splats; instead, it solely applied image loss $L_{img}$ for training. Compared to GaMeS \cite{waczynska2024games} and SuGaR \cite{guedon2023sugar}, our method enhances the PSNR by 2.0 dB and 0.2 dB, respectively.
    }
    \vspace{-10mm}
    \label{table:rendering-psnr-comparion-indoor}
\end{table}

\begin{table}[t]
    \centering
    \begin{adjustbox}{width=0.7\linewidth,center}
    \begin{tabular}{l|c|c|cc}
    \toprule
    \multirow{2}{*}{Method}& \multirow{2}{*}{Mesh} &\multirow{2}{*}{Rendering} &\multicolumn{2}{c}{Deep Blending \cite{hedman2018deep} } \\
    & & & DrJohnson & Playroom \\
    \midrule
    Instant-NGP \cite{muller2022instant} & X & Volume & \:\:27.7 / 0.84 / 0.38\:\:  & \:\:19.5 / 0.78 / 0.46\:\: \\
    mip-NeRF360 \cite{barron2022mip} & X & Volume & 29.1 / 0.90 / 0.24 & 29.6 / 0.90 / 0.25   \\
    3DGS \cite{kerbl20233d} & X & Splat & 29.1 / 0.90 / 0.24 & 30.0 / 0.90 / 0.24 \\
    \midrule
    MobileNeRF \cite{chen2023mobilenerf} & O & Mesh & 25.9 / 0.78 / 0.31  & 28.7 / 0.85 / 0.28  \\
    NeRF2Mesh \cite{tang2022nerf2mesh} & O & Mesh & 24.0 / 0.75 / 0.51 & 26.6 / 0.83 / 0.41   \\
    LTM \cite{Choi2024LTM} & O & Mesh & 26.5 / 0.82 / 0.40 & 29.1 / 0.87 / 0.32  \\
    SuGaR \cite{kerbl20233d} & O & Splat & 28.6 / 0.88 / 0.27 & 30.3 / 0.89 / 0.26 \\
    \midrule
     MeshGS (Ours) & O & Splat & \textbf{28.8}\:/\:\textbf{0.90}\:/\:\textbf{0.24} & \textbf{30.5}\:/\:\textbf{0.90}\:/\:\textbf{0.25}  \\
    \bottomrule
    \end{tabular}
    \end{adjustbox}
    \vspace{1mm}
    \caption{\textbf{Quantitative Comparison on Deep Blending Dataset.} We report PSNR \(\uparrow\), SSIM \(\uparrow\), and LPIPS \(\downarrow\) on Deep Blending datasets \cite{hedman2018deep}
    }
    \vspace{-5mm}
    \label{table:tandtdb}
\end{table}

\begin{figure}[h]
  \centering
    \includegraphics[width=0.97\linewidth]{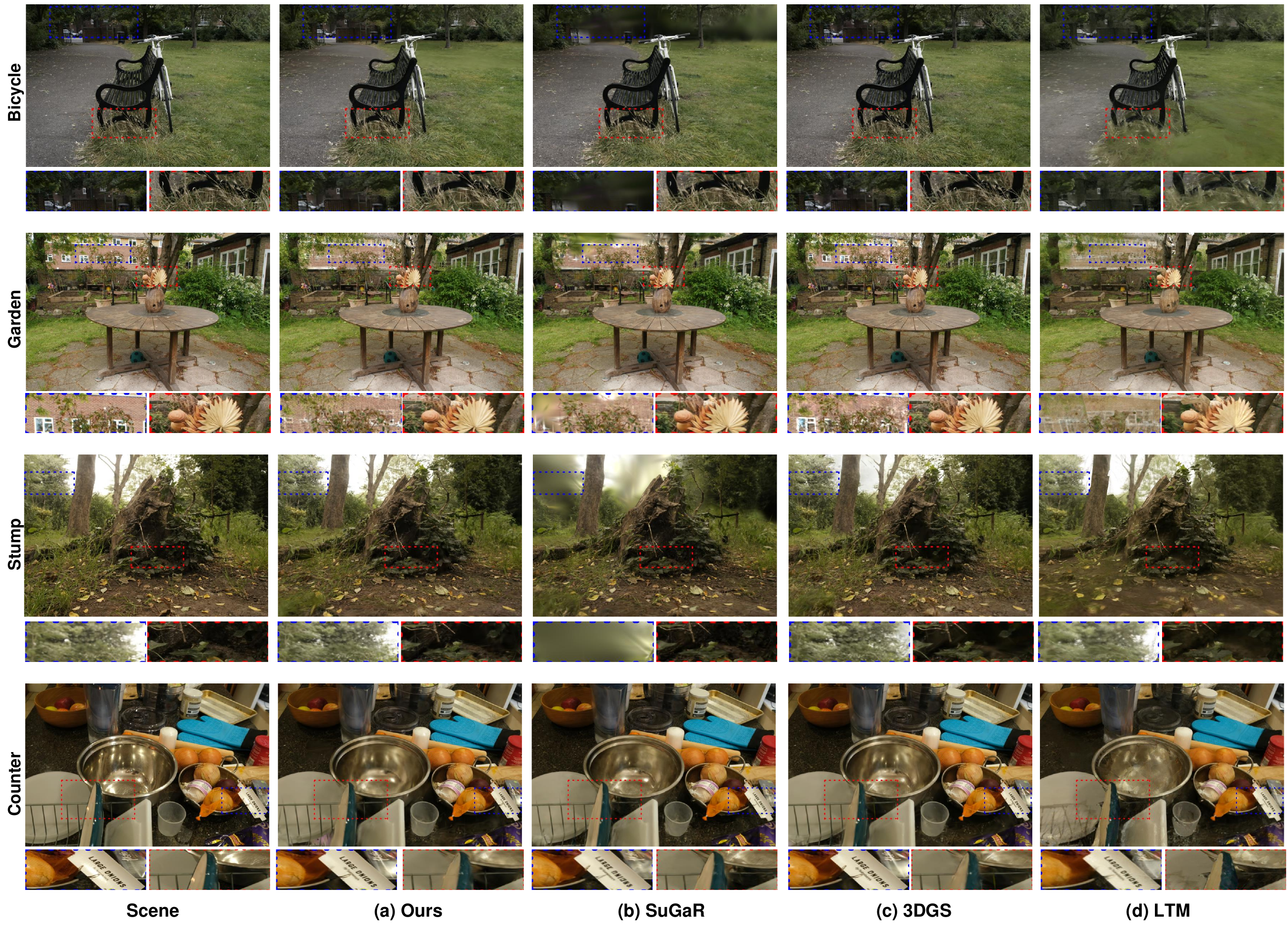}
    \vspace{-2mm}
   \caption{\textbf{Qualitative Comparisons with Existing Methods.} We visually compare our method with SuGaR \cite{guedon2023sugar}, 3DGS \cite{kerbl20233d}, and LTM \cite{Choi2024LTM}. For a detailed description, we visualize the zoomed-in blue and red regions.}
   \vspace{-5mm}
   \label{fig:render_comparison}
\end{figure}

\begin{figure}[h]
  \centering
    \includegraphics[width=0.99\linewidth]{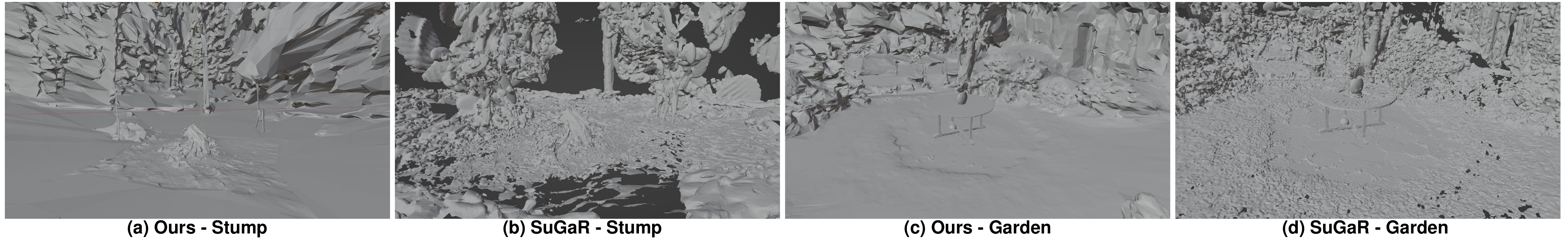}
    \vspace{-2mm}
   \caption{\textbf{Qualitative Comparisons of Our Method and SuGaR \cite{guedon2023sugar}}. We show shading mesh without texture.}
   \vspace{-4mm}
   \label{fig:sugar_ours_mesh}
\end{figure}

\begin{figure}[t]
    \centering
    \begin{minipage}[h]{0.48\textwidth}
    \includegraphics[width=0.99\linewidth]{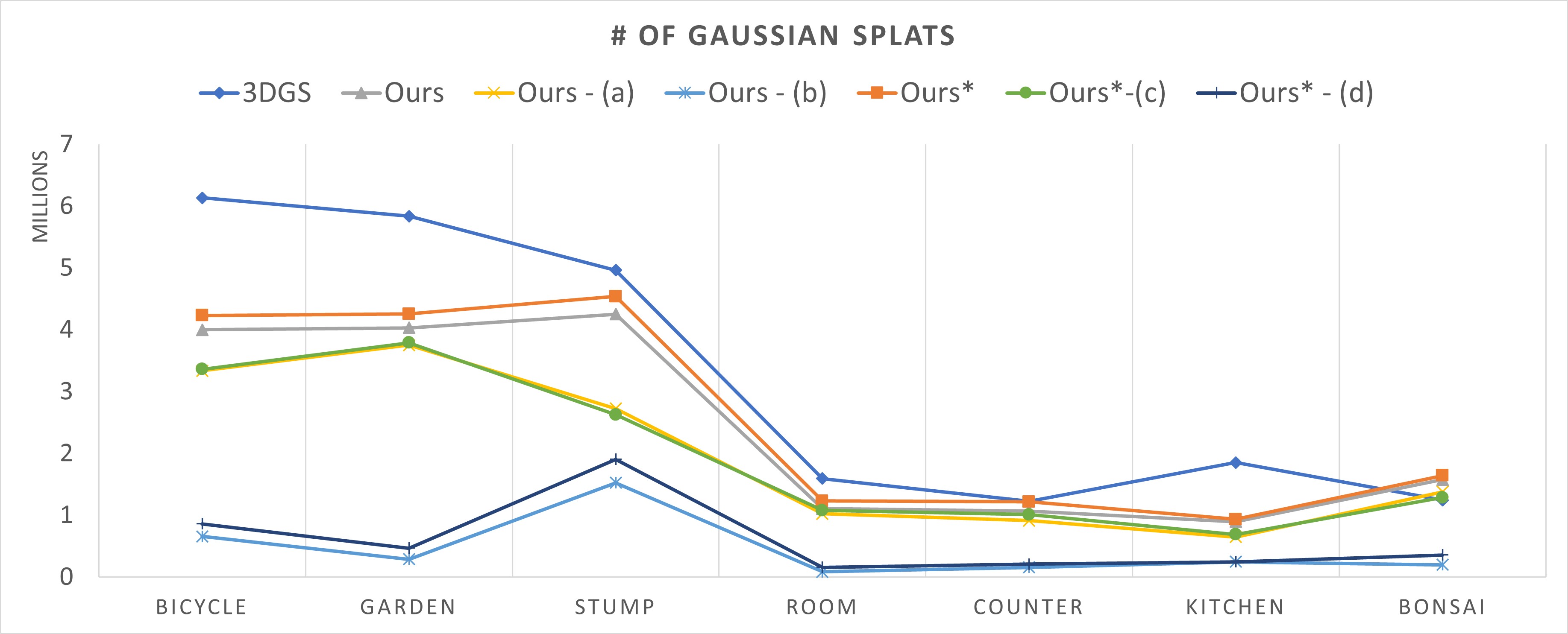} 
    \end{minipage}
    \vspace{-4mm}
    \begin{minipage}[t]{0.49\textwidth}
    \begin{adjustbox}{width=\linewidth}
    \begin{tabular}{l|ccccccc|c}
    \toprule
    \# of Splats & Bicycle & Garden & Stump & Room & Counter & Kitchen & Bonsai & Mean \\
    \midrule
    3DGS \cite{kerbl20233d} & 6132 & 5834 & 4961 & 1593 & 1222 & 1852 & 1244 &  3263 \\
    \midrule
    MeshGS & 3792 & 3838 & 4252 & 1103 & 1067 & 893 & 1576 & 2360  \\
    MeshGS-(a) (Tightly-bound Splats)  & 3236 & 3651 & 2723 & 1024 & 912 & 647 & 1378 & 1939 \\
    MeshGS-(b) (Loosely-bound Splats)  & 556 & 187 & 1522 & 82 & 156 & 245 & 198 & 421 \\
    \midrule
    MeshGS* & 4231 & 4255 & 4537 & 1230 & 1215 & 934 & 1640 & 2577  \\
    MeshGS*-(c) (Tightly-bound Splats)  & 3359 & 3788 & 2626 & 1077 & 1010 & 688 & 1283 & 1976 \\
    MeshGS*-(d) (Loosely-bound Splats)  & 859 & 467 & 1904 & 154 & 206 & 244 & 356 & 598 \\
    \bottomrule
    \end{tabular}
    \end{adjustbox}
    \end{minipage}
    \vspace{2mm}
    \caption{
    \textbf{The number of Gaussian Splats} The unit for the number of Gaussian splats is $10^3$. Compared to the original 3DGS, we utilize 30\% fewer Gaussian splats. MeshGS* did not utilize any regularization techniques to tightly align Gaussian splats; instead, it solely applied image loss $L_{img}$ for training. MeshGS-(a) and MeshGS*-(c) shows the number of tightly-bound Gaussian splats. MeshGS-(b) and MeshGS*-(d) denotes the number of loosely-bound Gaussian splats.  
    }
    \vspace{-2mm}
    \label{table:tight_loose_splats}
\end{figure}


\begin{figure}[h]
  \centering
    \includegraphics[width=0.99\linewidth]{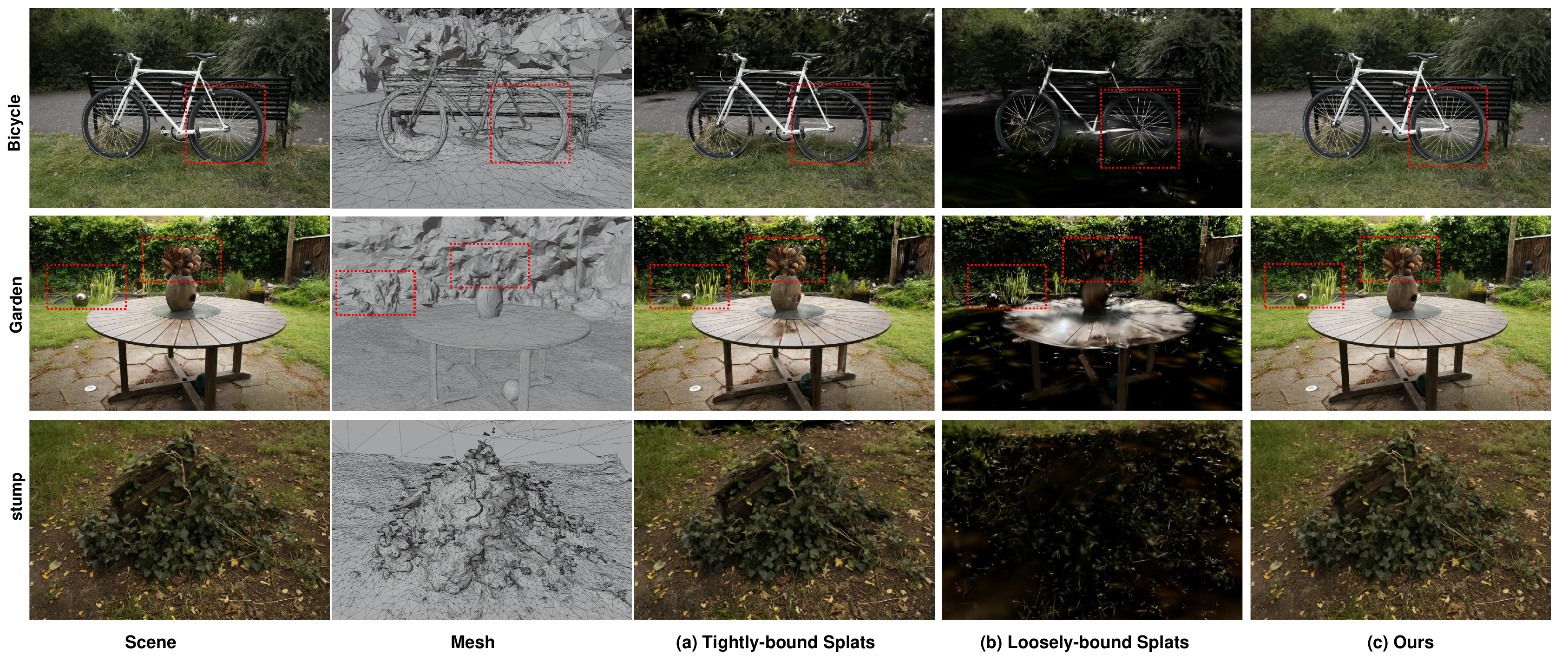}
    \vspace{-2mm}
   \caption{\textbf{Visualization results of both loosely-bound and tightly-bound Gaussian splats.} The second column shows wireframe mesh extracted without texture. The third column visualizes only the tightly-bound Gaussian splats, while the fourth column illustrates the loosely-bound Gaussian splats. The fifth column presents the final rendering results achieved by our method.}
   \vspace{-4mm}
   \label{fig:near_far_comparison}
\end{figure}

Furthermore, we train and evaluate on Playroom and DrJohnson scenes from the Deep Blending dataset \cite{hedman2018deep}. These two scenes are collected from bounded indoor scenes and have many complex textures on Playroom scene and contain many textureless regions on DrJohnson scene. 
These two scenes were captured from bounded indoor environments, with the Playroom scene consisting intricate textures and the DrJohnson scene comprising several textureless regions.
In Table \ref{table:tandtdb}, Our method demonstrates comparable rendering quality to both the original 3DGS \cite{kerbl20233d} and SuGaR \cite{guedon2023sugar}.

In Fig. \ref{fig:render_comparison}, we visualize the rendering results for comparison. Our method builds on the mesh reconstructed by LTM \cite{Choi2024LTM}. Compared to LTM, the Gaussian splatting in our method exhibits sharper and more high-frequency details than the neural appearance field. As meshes struggle to represent highly detailed geometry such as dense foliage, tree leaves, and thin iron structures, LTM may not adequately capture these appearances for rendering (see the zoomed-in regions in Fig. \ref{fig:render_comparison}). SuGaR is based on the original 3DGS and reconstruct the mesh solely based on Gaussian splats. However, compared to our geometry reconstruction, the mesh generated by SuGaR exhibits numerous geometric artifacts, particularly in the background region, where defects are prevalent due to weak smoothness regularization (see the background region in Fig. \ref{fig:mesh_comparison} and Fig. \ref{fig:sugar_ours_mesh}). Due to the geometric artifacts present in the background region, SuGaR often produces blurry rendering results, especially in unbounded outdoor scenes (see the background regions of Bicycle, Garden, and Stump in Fig. \ref{fig:render_comparison}).

\subsection{Ablation Study}
In Table \ref{table:rendering-psnr-comparion-outdoor} and Table \ref{table:rendering-psnr-comparion-indoor}, MeshGS* (ours) denotes we only use image loss $L_{img}$ to train our mesh-based Gaussian splatting in Section \ref{sec:Training}. Compared to MeshGS (ours), MeshGS* did not employ any regularization technique to tightly align Gaussian splats with the mesh surface. We observe that MeshGS* shows slightly better rendering quality compared to our method. This is a trade-off between aligning splats geometrically and achieving high-quality rendering.  

In Table \ref{table:tight_loose_splats},  we show the proportion of tightly-bound Guassian splats and loosely-bound Gaussian splats. Compared to MeshGS*, our approach employs fewer Gaussian splats overall. Furthermore, our method utilizes Gaussian splats that are 6\% less tightly bound compared to MeshGS*. It is noteworthy that the Stump scene generates a significantly higher proportion of loosely-bound Gaussian splats (42\%) compared to other scenes (22\%). In the Stump scene, we observe that the majority of background regions are covered by loosely-bound Gaussian splats. This is due to the inability of the mesh to accurately cover all background regions, causing Gaussian splats to often move far from the mesh surface.

Figure \ref{fig:near_far_comparison} presents a visualization of both loosely-bound and tightly-bound Gaussian splats using the outdoor mip-NeRF 360 dataset. In Fig. \ref{fig:near_far_comparison}-(a), tightly-bound Gaussian splats can cover the most regions for rendering. Nevertheless, due to limitations in mesh geometry, they fail to address all mesh artifacts. In Fig. \ref{fig:near_far_comparison}, the red dotted box highlights regions where tightly-bound Gaussian splats cannot accurately represent details such as the bicycle wheel, thin trunk around the vase, and dense tall grass, which closely resemble mesh artifacts. Loosely-bound Gaussian splats can serve as a supplementary method to cover these artifacts. In the Stump scene, loosely-bound Gaussian splats demonstrate their ability to accurately depict the detailed thin grass and leaves covering the trunk, thereby facilitating high-fidelity rendering. Moreover, given that the mesh geometry shows more geometric defects in the background region, it necessitates the use of numerous loosely-bound Gaussian splats to achieve high-quality rendering. Furthermore, in the mesh geometry, the process of mesh decimation significantly reduces the number of triangles in the background region. As a result, the initialization of Gaussian splats becomes more sparse, posing challenges in rendering images.

\begin{figure}[h]
  \centering
    \includegraphics[width=0.8\linewidth]{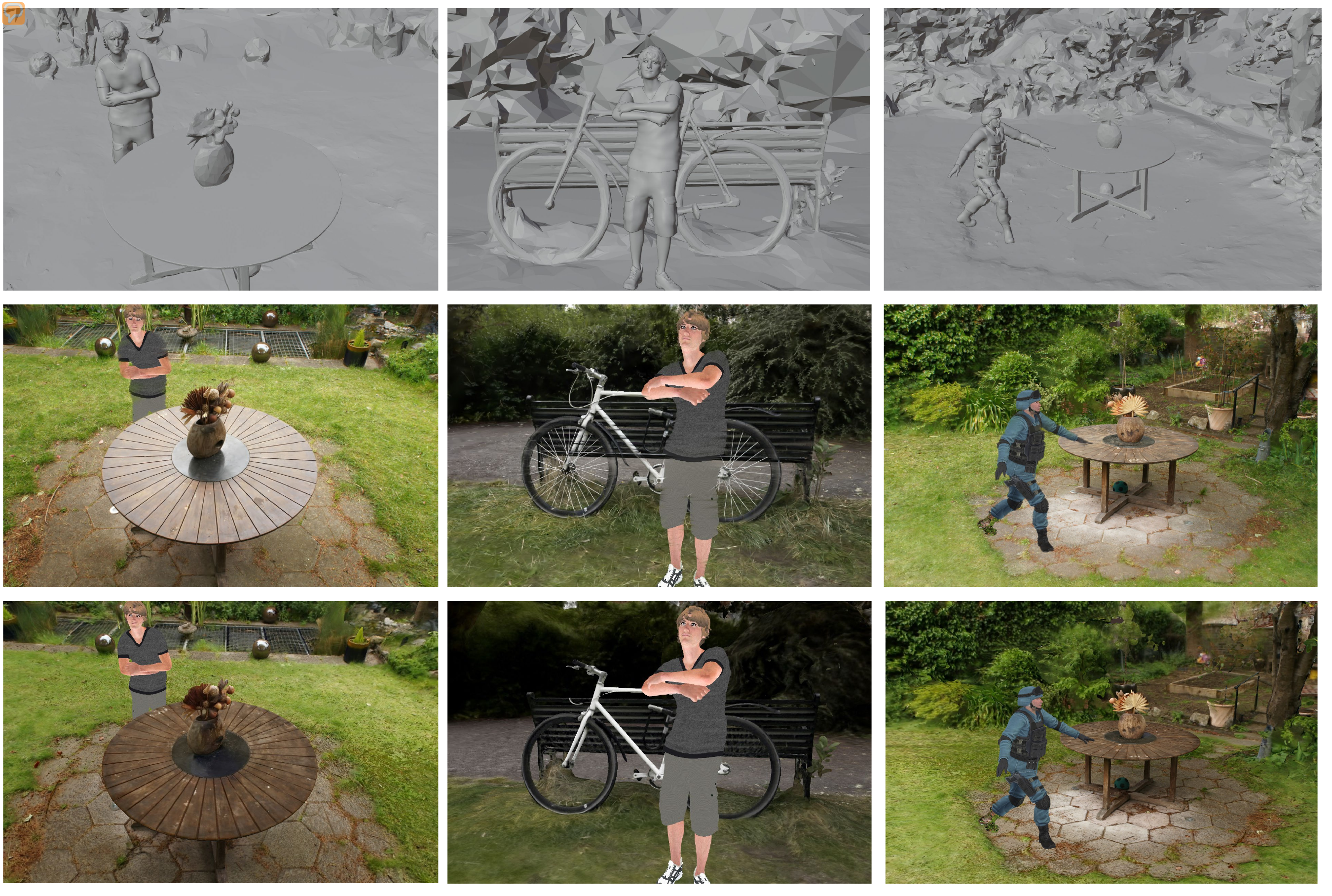}
    \vspace{-3mm}
   \caption{\textbf{Scene Composition}: Rendering obtained by embedding the synthetic human to a reconstructed scene. The first row visualizes the shading mesh without texture. The second row shows the composition of a human in the scene reconstructed by our method. The third row visualizes the composition of the same human in the scene generated by LTM \cite{Choi2024LTM}.}
   \vspace{-5mm}
   \label{fig:composition}
\end{figure}
\subsection{Compositional scenes}
In Fig. \ref{fig:composition}, we visualize the practical application of our method for scene composition. We incorporate an artist-designed human model from Mixamo \cite{Mixamo} into the scene reconstructed by our method. This human asset consists of high-polygon meshes with photorealistic materials and textures. We show the composition of a synthetic human into a different scene. The background region rendered by our method shows photorealistic image quality of compositional scenes.  Thanks to the mesh, we can composite the object and scene. Furthermore, the composited scene can be rendered using Gaussian splatting without causing severe artifacts around the human's boundary.

\section{Conclusion, Limitations, and Future Work}
\label{sec:conclusion}
This paper presents a practical approach for integrating mesh representation with 3D Gaussian splatting rendering. our method attains a high level of photorealistic rendering quality by employing Gaussian splats bound to the mesh surface and executing distance-based Gaussian splatting to ensure alignment with the mesh surface. Additionally, we differentiate between two types of Gaussian splats and present training strategies tailored to their respective roles. We present that a combination of mesh-based representation and Gaussian splatting is crucial for enhancing rendering quality and ensuring compatibility with game engines in practical applications.   However, our method does have limitations. Compared to textured mesh, our approach necessitates large disk storage and a high-end GPU for real-time rendering. Moreover, we require a more sophisticated method for assessing the quality of Gaussian splats and ensuring their tight binding to the surface mesh. Additionally, our method is constrained by the underlying mesh shape, particularly when dealing with excessively large faces that cannot appropriately initialize Gaussian splats. In the future, our goal is to develop surface extraction techniques from loosely-bound splats to address mesh defects and evaluate geometric quality. Furthermore, we should explore adaptive methods for distributing Gaussian splats to individual triangles, akin to mesh subdivision or remeshing techniques. 

\textbf{Acknowledgements} 
This work was supported in part by ARO Grant W911N
F2310352 and Army Cooperative Agreement W911NF2120076.  

\appendix

\section{Implementation Details}
We utilize the geometry reconstruction and mesh processing part from the LTM \cite{Choi2024LTM}. Leveraging the strong smoothness regularization \cite{eikonal}, this method shows consistently smooth surfaces on both ground and wall structures. Additionally, due to the incorporation of contraction function \cite{yariv2023bakedsdf}, background regions are effectively reconstructed. However, we do not utilize the optimization step, resulting in a mesh surface that lacks detailed structures and often appears smoothed due to the limited number of vertex. Our method is implemented using PyTorch \cite{PyTorch} and builds on Gaussian rasterization provided by the original 3DGS \cite{kerbl20233d}. For scene composition, we initially obtain the textured human mesh from Mixamo and export the compositional mesh result from the Blender engine. Subsequently, we employ our rasterization code to render the compositional scene using the exported mesh composition.


\section{Comparison to 2D Gaussian Splatting}
Recently, a 2DGS \cite{huang20242d} was released, showing remarkable performance in mesh reconstruction and novel view synthesis. In Table \ref{table:rendering-psnr-comparion-outdoor} and Table \ref{table:rendering-psnr-comparion-indoor}, we compare our method with 2D GS in terms of rendering performance. Unlike approaches that bind 3D splats to an underlying mesh, this research prioritizes aligning 2D Gaussian splats with the underlying geometry without requiring prior 3D geometry. This paper demonstrate that 2D Gaussian splats surpass 3D Gaussian splats in representing geometry. Our Gaussian splats binding method can leverage the 2D Gaussian splats representation to bind splats to a given mesh.

\begin{table}[h]
    \centering
    \begin{adjustbox}{width=\linewidth,center}
    \begin{tabular}{l|c|c|cccc}
    \toprule
    \multirow{2}{*}{Method}&\multirow{2}{*}{Mesh}&\multirow{2}{*}{Rendering}&\multicolumn{4}{c}{Outdoor (PSNR \(\uparrow\)  / SSIM \(\uparrow\)  / LPIPS \(\downarrow\)) } \\
    & & & Bicycle & Garden & Stump & Mean \\
    \midrule
    2DGS \cite{huang20242d} & O & Splat & \underline{24.6} / 0.71 / 0.30 & 26.6 / 0.83 / 0.16 & \underline{26.0} / \underline{0.76} / 0.29 & 25.7 / 0.76 / 0.25  \\
    MeshGS (Ours) & O & Splat & 24.4 / \underline{0.71} / \underline{0.26} & \underline{26.8} / \underline{0.85} / \underline{0.12} & 25.8 / 0.74 / \underline{0.25} & \underline{25.7} / \underline{0.77} / \underline{0.21}  \\
    MeshGS* (Ours) & O & Splat & \textbf{24.9}\:/\:\textbf{0.75}\:/\:\textbf{0.25} & \textbf{26.8}\:/\:\textbf{0.86}\:/\:\textbf{0.12} & \textbf{26.3}\:/\:\textbf{0.76}\:/\:\textbf{0.22} & \textbf{26.0}\:/\:\textbf{0.79}\:/\:\textbf{0.20} \\
    \bottomrule
    \end{tabular}
    \end{adjustbox}    
    \vspace{1mm}
    \caption{\textbf{Quantitative Comparison on outdoor mip-NeRF 360 Dataset.} MeshGS* did not utilize any regularization techniques to tightly align Gaussian splats; instead, it solely applied image loss for training.   
    }
    \vspace{-3mm}
    \label{table:rendering-psnr-comparion-outdoor}
\end{table}

\begin{table}[h]
    \centering
    \begin{adjustbox}{width=\linewidth,center}
    \begin{tabular}{l|c|c|cccccc}
    \toprule
    \multirow{2}{*}{Method}&\multirow{2}{*}{Mesh}&\multirow{2}{*}{Rendering}&\multicolumn{5}{c}{Indoor (PSNR \(\uparrow\)  / SSIM \(\uparrow\)  / LPIPS \(\downarrow\))}\\ 
    & & & Room & Counter & Kitchen & Bonsai & Mean \\
    \midrule
    2DGS \cite{huang20242d} & O & Splat & \underline{30.8} / \underline{0.91} / \textbf{0.22} & \textbf{28.1} / 0.89 / 0.22 & \textbf{30.1} / \textbf{0.92} / \textbf{0.14} & \textbf{31.3} / 0.93 / 0.25 & \textbf{30.1} / 0.91 / \textbf{0.19}   \\
    MeshGS (Ours) & O & Splat & 30.3 / 0.90 / 0.24 & \underline{28.0} / \underline{0.89} / \textbf{0.21}  & \underline{29.7}\:/\: 0.90\:/\:\underline{0.17}  & \underline{30.7}\:/\:\textbf{0.93} / \underline{0.20} & \underline{29.6}\:/\:\textbf{0.91}\:/\:\underline{0.20} \\
    MeshGS* (Ours) & O & Splat & \textbf{30.9}\:/\:\textbf{0.93}\:/\:\underline{0.23} & 27.8 / \textbf{0.90} / \underline{0.21} & 29.3 / \underline{0.90} / 0.17 & 30.5 / \underline{0.93} / \textbf{0.19} & 29.6 / 0.91 / 0.20  \\
    \bottomrule
    \end{tabular}
    \end{adjustbox}
    \vspace{1mm}
    \caption{\textbf{Quantitative Comparison on indoor mip-NeRF 360 Dataset.} MeshGS* did not utilize any regularization techniques to tightly align Gaussian splats; instead, it solely applied image loss for training.   
    }
    \vspace{-10mm}
    \label{table:rendering-psnr-comparion-indoor}
\end{table}

\section{Additional Qualitative Comparisons}
\begin{figure}[h]
  \centering
    \includegraphics[width=0.95\linewidth]{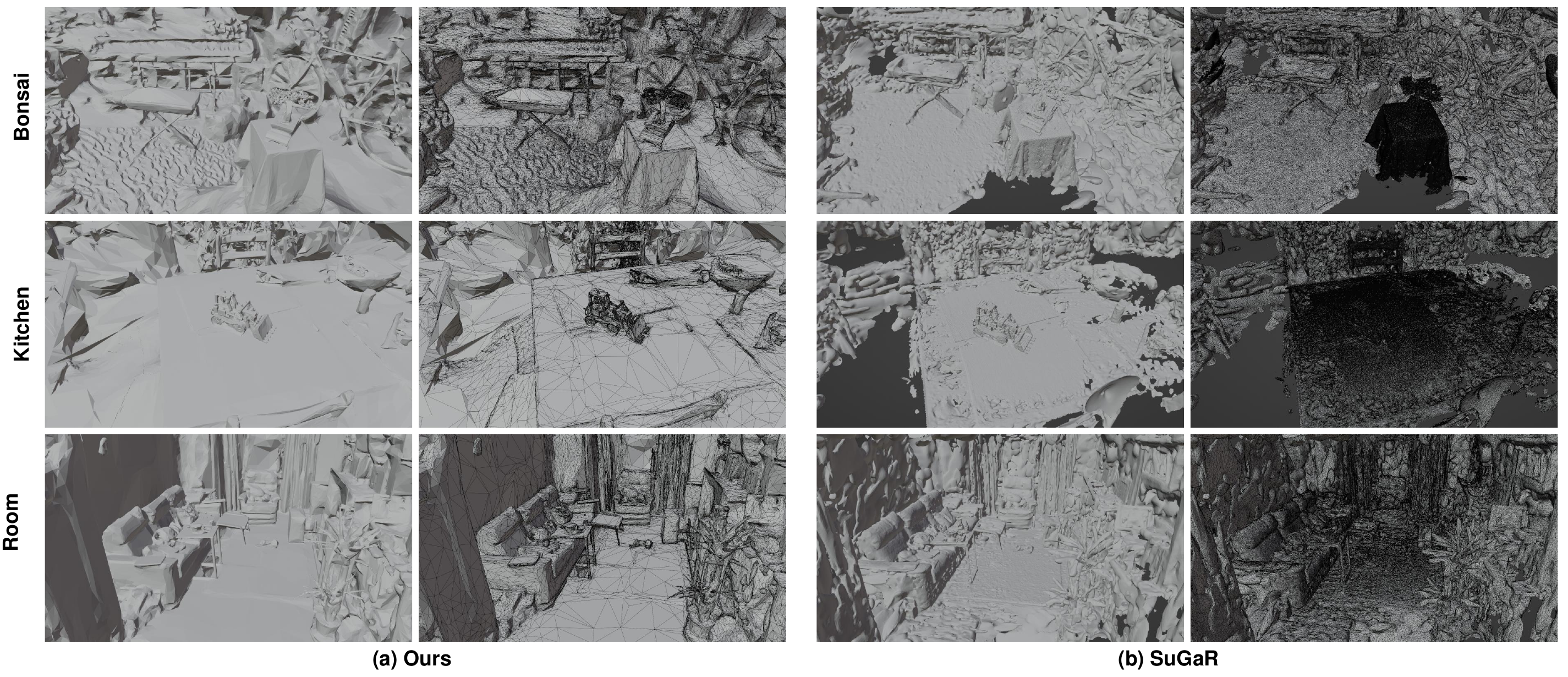}
    \vspace{-2mm}
   \caption{ Qualitative comparison between (a) Ours and (b) SuGaR \cite{guedon2023sugar}. Our method and SuGaR show shading mesh and wireframe mesh extracted without texture, respectively. In SuGaR, the dark grey regions denote empty areas where the mesh fails to represent the scene. Compared to ours, SuGaR's mesh inadequately captures background regions}
   \label{fig:mesh_comparison_mip360}
\end{figure}

\begin{figure}[h]
  \centering
    \includegraphics[width=0.95\linewidth]{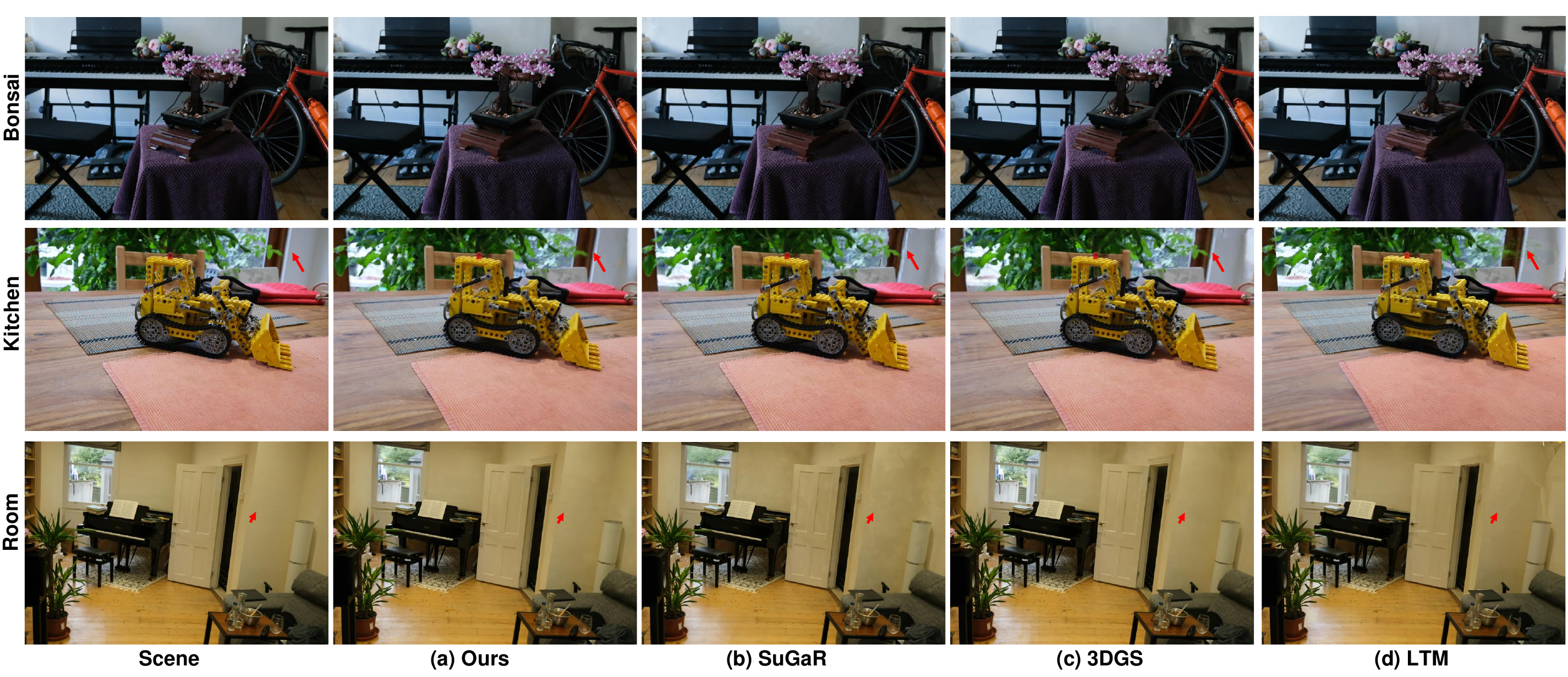}
    \vspace{-2mm}
   \caption{\textbf{Qualitative Comparisons with Existing Methods.} We visually compare our method with SuGaR \cite{guedon2023sugar}, 3DGS \cite{kerbl20233d}, and LTM \cite{Choi2024LTM}. Red arrows emphasize subtle differences in rendering quality}
   \label{fig:render_comparison_mip360}
\end{figure}

\begin{figure}[h]
  \centering
    \includegraphics[width=0.95\linewidth]{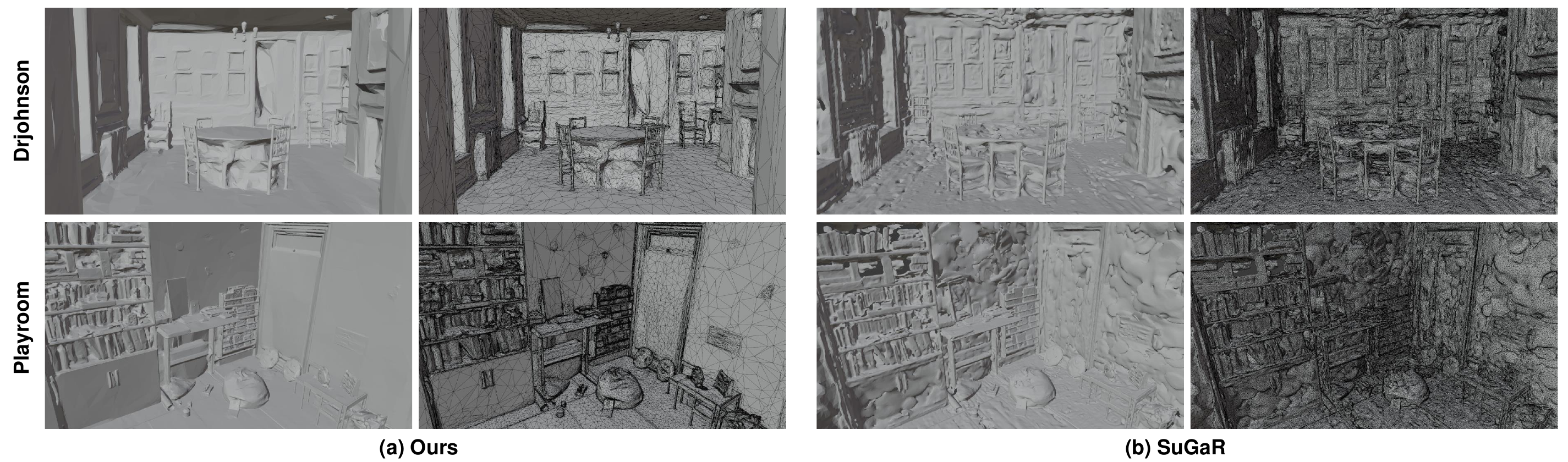}
   \caption{Qualitative comparison between (a) Ours and (b) SuGaR \cite{guedon2023sugar}. Our method and SuGaR show shading mesh and wireframe mesh extracted without texture, respectively.}
   \vspace{-3mm}
   \label{fig:mesh_comparison_db}
\end{figure}

\begin{figure}[h]
  \centering
    \includegraphics[width=0.95\linewidth]{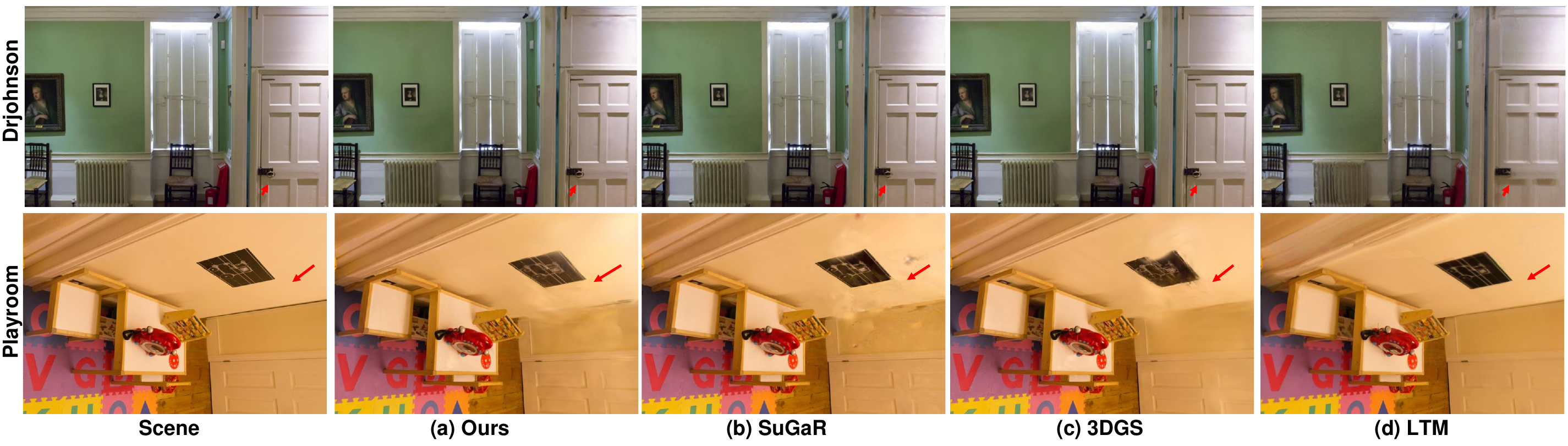}
   \caption{\textbf{Qualitative Comparisons with Existing Methods.} We visually compare our method with SuGaR \cite{guedon2023sugar}, 3DGS \cite{kerbl20233d}, and LTM \cite{Choi2024LTM}. Red arrows emphasize subtle differences in rendering quality.}
   \vspace{-3mm}
   \label{fig:render_comparison_db}
\end{figure}

Supplementary material includes all rendering results on the test set, provided as a downloadable zip file. To accommodate file size limitations (100mb for supplementray material), we downscale the resolution of rendering images. In the manuscript, we show the mesh reconstruction results for Garden, Stump, and Bicycle scenes (refer to Fig. 2, Fig.4, and Fig. 6) due to the limited space. we visualize the results for the remaining scenes on both Mip-NeRF 360 dataset and Deep Blending dataset. We present both the mesh reconstruction and rendering outcomes for these scenes, depicted in Fig. \ref{fig:mesh_comparison_mip360} and Fig. \ref{fig:render_comparison_mip360}, respectively. Additionally, we show visualizations for the mesh reconstruction and rendering results of the Playroom and Dr. Johnson scenes from the Deep Blending dataset, showcased in Fig. \ref{fig:mesh_comparison_db} and Fig. \ref{fig:render_comparison_db}, respectively. 

\vspace{-5mm}
\vspace{-15mm}

\newpage

%
%
\bibliographystyle{splncs04}
\bibliography{main}
\end{document}